\documentclass[sigconf]{acmart}

\usepackage{multirow}
\usepackage{balance}

\AtBeginDocument{%
  \providecommand\BibTeX{{%
    \normalfont B\kern-0.5em{\scshape i\kern-0.25em b}\kern-0.8em\TeX}}}

\copyrightyear{2021} 
\acmYear{2021} 
\setcopyright{acmcopyright}\acmConference[MM '21]{Proceedings of the 29th ACM International Conference on Multimedia}{October 20--24, 2021}{Virtual Event, China}
\acmBooktitle{Proceedings of the 29th ACM International Conference on Multimedia (MM '21), October 20--24, 2021, Virtual Event, China}
\acmPrice{15.00}
\acmDOI{10.1145/3474085.3475301}
\acmISBN{978-1-4503-8651-7/21/10}

\begin{document}
\fancyhead{}

\title{Learning Segment Similarity and Alignment in Large-Scale Content Based Video Retrieval}

\author{Chen Jiang}
\authornote{These authors contributed equally to this research.}
\email{qichen.jc@antgroup.com}

\author{Kaiming Huang}
\authornotemark[1]
\email{kaiming.huangkm@antgroup.com}
\affiliation{%
  \institution{Ant Group}
  \country{}
}

\author{Sifeng He}
\authornotemark[1]
\email{sifeng.hsf@antgroup.com}

\author{Xudong Yang}
\authornotemark[1]
\email{jiegang.yxd@antgroup.com}
\affiliation{%
  \institution{Ant Group}
  \country{}
}

\author{Wei Zhang}
\authornote{Corresponding author.}
\email{ivy.zw@antgroup.com}

\author{Xiaobo Zhang}
\email{ayou.zxb@antgroup.com}
\affiliation{%
  \institution{Ant Group}
  \country{}
}

\author{Yuan Cheng}
\email{chengyuan.c@antgroup.com}

\author{Lei Yang}
\email{yl149505@antgroup.com}
\affiliation{%
  \institution{Ant Group}
  \country{}
}

\author{Qing Wang}
\email{wq176625@antgroup.com}

\author{Furong Xu}
\email{booyoungxu.xfr@antgroup.com}
\affiliation{%
  \institution{Ant Group}
  \country{}
}

\author{Tan Pan}
\email{pantan.pt@antgroup.com}

\author{Wei Chu}
\email{weichu.cw@antgroup.com}
\affiliation{%
  \institution{Ant Group}
  \country{}
}


\renewcommand{\shortauthors}{Jiang,Huang,He and Yang, et al.}

\begin{abstract}
  With the explosive growth of web videos in recent years, large-scale Content-Based Video Retrieval (CBVR) becomes increasingly essential in video filtering, recommendation, and copyright protection. Segment-level CBVR (S-CBVR) locates the start and end time of similar segments in finer granularity, which is beneficial for user browsing efficiency and infringement detection especially in long video scenarios. The challenge of S-CBVR task is how to achieve high temporal alignment accuracy with efficient computation and low storage consumption. 
    In this paper, we propose a Segment Similarity and Alignment Network (SSAN) in dealing with the challenge which is firstly trained end-to-end in S-CBVR. SSAN is based on two newly proposed modules in video retrieval: (1) An efficient Self-supervised Keyframe Extraction (SKE) module to reduce redundant frame features, (2) A robust Similarity Pattern Detection (SPD) module for temporal alignment. 
    In comparison with uniform frame extraction, SKE not only saves feature storage and search time, but also introduces comparable accuracy and limited extra computation time. In terms of temporal alignment, SPD localizes similar segments with higher accuracy and efficiency than existing deep learning methods. Furthermore, we jointly train SSAN with SKE and SPD and achieve an end-to-end improvement. Meanwhile, the two key modules SKE and SPD can also be effectively inserted into other video retrieval pipelines and gain considerable performance improvements.
    Experimental results on public datasets show that SSAN can obtain higher alignment accuracy while saving storage and online query computational cost compared to existing methods.
    
   %
\end{abstract}

\ccsdesc[500]{Information systems~Video search}
\ccsdesc[500]{Information systems~Near-duplicate and plagiarism detection}
\ccsdesc[500]{Information systems~Similarity measures}

\keywords{Large-Scale Segment-level Content Based Video Retrieval, Temporal Alignment, Keyframe Extraction, Similarity Pattern Detection}



\maketitle

\section{Introduction}
With the rapid development of video-sharing services like TikTok, Youtube, etc., a massive amount of videos are generated. Consequently,
it is essential both for users to search and access videos efficiently and for service platforms to protect copyright and remove duplicate or low-quality videos.
Hence, content-based video retrieval (CBVR) system is important in applications like video filtering, recommendation, management, and copyright protection.
In detail, given a query video and a database as search range, a CBVR system retrieves videos with similar content as the query from the database with limited system response time.
In scenarios of copyright detection, video filtering, etc., a finer granularity of retrieval is further required from video level to segment level for the precise location. It is time consuming to scan all retrieved videos from start to end to determine which clip is relevant to the query, especially when video length is long.
Therefore, it is necessary to develop a Segment-level CBVR (S-CBVR) system that both retrieves video segments and temporally aligns similar segments.

 Considering the huge quantity of videos,an S-CBVR system needs to achieve high retrieval and temporal alignment performance at low storage and computation requirements.
S-CBVR in large-scale case
is a challenging task.
Firstly, there are a massive amount of videos to be retrieved and thus the cost of resources for computation and storage is heavy.
For example, 2 billion Youtube users watch about 1 billion hours of video content every day\footnote{https://www.youtube.com/yt/about/press/, accessed 11 Nov 2020}. 
Secondly, there may exist a gap on both the visual appearance and temporal sequence between the query video and related gallery videos due to shooting condition variations, or various user editing techniques. The variations of similar videos may be larger than dissimilar ones. To address these issues, we introduce SSAN, a Segment Similarity and Alignment Network that is suitable for large-scale segment-level video retrieval. A sample query process of SSAN is illustrated in Figure 1. Different from the common methods ~\cite{kordopatis2019visil,feng2018video,Poullot2015,LAMVBaraldi2018} with video pairs as input for one on one comparison, we adopt high dimensional index based retrieval which computes features of database videos offline and building an index to accelerate the online query. To ensure the generalization ability for various transformations, fine-grained frame-level features 
are adopted. 

\begin{figure*}
\begin{center}
\includegraphics[width=14cm,keepaspectratio]{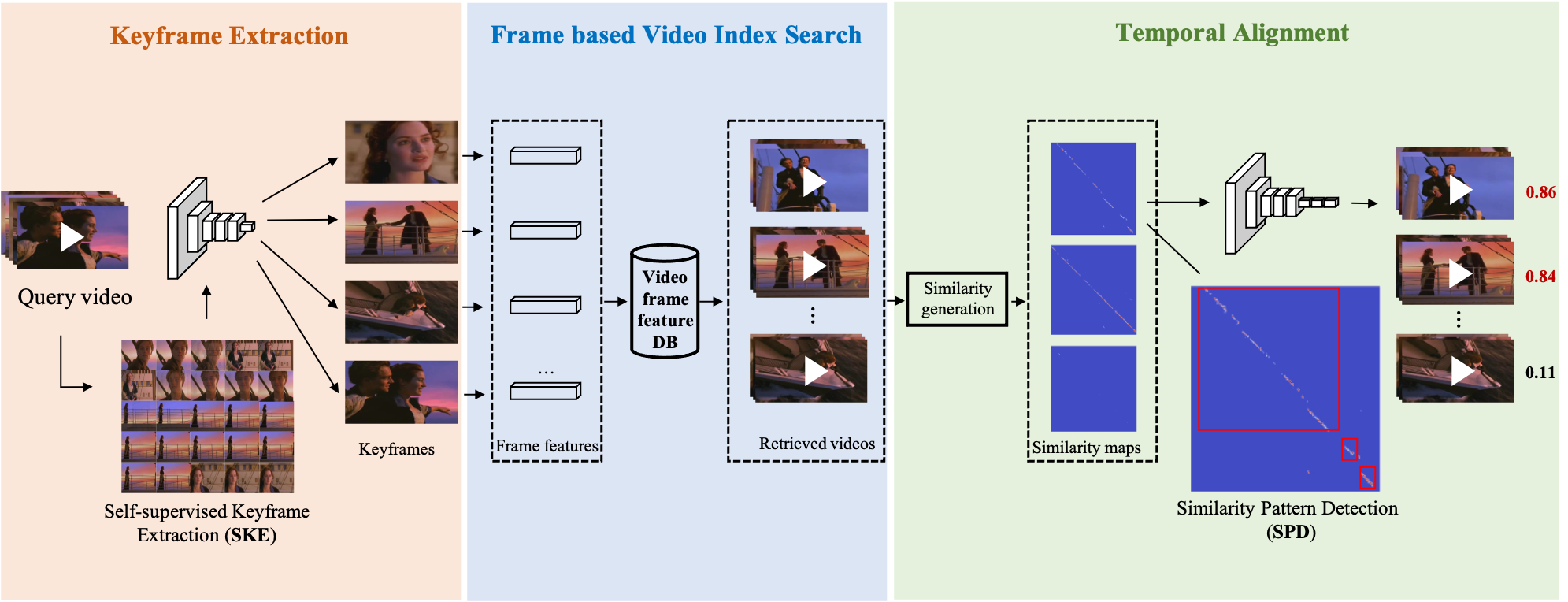}
\end{center}
   \caption{Query process of our proposed approach on Segment-level Content Based Video Retrieval (S-CBVR)}
\label{fig:short}
\end{figure*}
As for frame extraction, existing approaches usually extract frames with uniform time interval, which may bring many redundant and low-quality features.
For instance, in news broadcast videos with almost unchanging pictures, features are extracted redundantly with almost the same information.
Moreover, low-quality frames with problems like motion blur, out of focus, or stripe noise, etc., may be extracted. 
Here, we propose a Self-supervised Keyframe Extraction (SKE) method to reduce redundant frames and improve frame quality.
SKE is a lightweight network and uses annotations generated from a teacher model which can extract high-quality frames. This avoids subjective human annotations. 

A temporal alignment module needs to reveal the similarity and time range of infringing segments between query and gallery videos. Traditional alignment approaches adopts Temporal Hough Voting~\cite{voting1,vcdb}, Dynamic Programming~\cite{dp1} or Temporal Networks~\cite{temporaNetwork} to accomplish temporal alignment.
These methods require regular similarity patterns and are not robust to capture diverse patterns. 
Recent deep learning methods using Fourier-domain representations\cite{LAMVBaraldi2018} achieves better performance but is still time consuming.
We propose SPD, which formulates temporal alignment as a pattern detection task on similarity maps produced from frame-level features of query and gallery videos. The detection task is trained on a lightweight network, which can learn a large variety of similarity patterns from plentiful data and thus is robust.


Furthermore, we propose SSAN  with SPD and SKE and train end-to-end to gain further improvements. The SKE learns keyframe confidence scores of uniform frames, which are then fused with similarity maps and fed to SPD. The pairwise frames with high keyframe confidence scores make a difference to detect similar segments in SPD.
During training, similarity scores are pre-computed, and the detection loss of SPD is propagated to SKE to guide a better keyframe selection. During query shown in Figure 1, the keyframe confidence scores are quantized and the similarity score is computed based on index search results to facilitate efficient large-scale retrieval. 

To our best knowledge, the end-to-end network and training process are firstly proposed in large-scale video retrieval. The STRNN \cite{Hu_2019} claims an end-to-end framework but only contains the temporal alignment part as our SPD, which excludes the keyframe extraction part. 
Our work can bring a new insight on how to train a video similarity network in large-scale scenarios using indexing techniques. 

\section{Related Work}
In CBVR area, similarity definition varies in different tasks. In Video Copy Detection (VCD), videos that originate from the source video are regarded as similar. In Near Duplicate Video Retrieval(NDVR), videos captured from the same scenario are defined as similar, while in Fine-grained Incident Video Retrieval (FIVR)\cite{kordopatis2019fivr},videos that describe the same event are similar. Video Re-localization \cite{videoRel2018} finds similar event locations in video pairs. 
The video similarity calculation methods can be classified, based on feature extraction and similarity calculation process employed, 
in two categories: online one vs one video matching, and offline index-based retrieval. 
Specifically, video matching methods input two videos into a network to calculate similarity \cite{kordopatis2019visil} or locate similar segments \cite{feng2018video,Poullot2015}. This is not feasible in large-scale situations in terms of the unacceptable response time. 
To find a user-interested video clip from a given large-scale video database, index-based methods
extract features of database videos offline and build index of them. Then the online query is executed on the index, which enables millions or even billions of videos to be searched in a proper time. 

In this section, we will give an overview of some fundamental works that have contributed to our work. It can be mainly divided into three parts: feature extraction and indexing, keyframe extraction and temporal alignment.
\subsection{Feature Representation and Indexing}
Frame-level features are proved to gain a large margin in retrieval tasks \cite{kordopatis2019visil,Shao2020ContextEF} and are necessary to locate similar segments.
Current methods employ Deep Convolutional Neural Networks (CNNs) \cite{mac17} and Deep Metric Learning (DML) \cite{ml2017,Shao2020ContextEF} to extract robust features. The application of Maximum Activation of Convolutions (MAC) and its variants \cite{mac17,mac2} 
has proved to be an efficient representation in retrieval tasks. Additionally, some recent works rely on contrastive loss \cite{Shao2020ContextEF} or triplet loss \cite{xu2020metric} to learn better video representations. 

Indexing techniques play an import role in large-scale S-CBVR system. Product Quantization \cite{pq} quantity features into clustering centers to accelerate the access to similar features. Deep hash \cite{deephash1,deephash2} transforms features into binary codes for fast Hamming distance computation. Hierarchical Navigable Small World (HNSW) \cite{hnsw} constructs a neighborhood relationship graph offline to search fast online. Search results include similarities between frames and their corresponding timestamps. 
Then similarity matrix, temporal network, etc., based on frame-level similarities, can be constructed to generate segment retrieval results.  

In this paper, we focus on practical retrieval task based on indexing scheme in large-scale datasets, which requires both high retrieval performance and high computational efficiency. 
Particularly, the non-differentiability of the indexing operation leads to difficulties in the end-to-end video similarity learning.

\subsection{Keyframe Extraction}
Most S-CBVR  methods extract frames uniformly considering the simplicity in temporal alignment, especially for traditional non deep learning methods\cite{vcdb,voting1,dp1,temporaNetwork}. As mentioned above, extracting frames uniformly results in redundant features. Thus, we introduce keyframe extraction to reduce redundancy and avoid low-quality frames.
Early methods formulate keyframe extraction as a clustering problem based on frame-level handcrafted features including color, texture, SIFT, etc \cite{Avila2011VSUMMAM,vscanMohamedIG14}. More recent works \cite{jadon2019video,zhou2017reinforcevsumm,cnnsummary2018,zhang2016lstm,sumgan2017} rely on  deep CNN features combined with aggregation methods like clustering\cite{jadon2019video}, Long Short-Term Memory (LSTM)\cite{zhang2016lstm}, reinforcement learning\cite{zhou2017reinforcevsumm}, Generative Adversarial Networks(GANs)\cite{sumgan2017}.
In \cite{2018fcnsum,hand2020}, keyframe extraction is addressed as a segmentation problem and multiple frames are organized into one picture to locate keyframes. Compared to \cite{hand2020}, our SKE is formulated as a multi-frame classification problem. 
Meanwhile, due to the subjectivity and difficulty caused by human annotations, the work \cite{hand2020} adopted VSUMM algorithms \cite{Avila2011VSUMMAM}to generate annotations to train a lightweight model. Additionally, \cite{optimalgroup2016} proposed a method for temporal grouping of scenes, which formulated the problem as an optimal sequential grouping based on a distance matrix. Inspired by \cite{hand2020} and \cite{optimalgroup2016}, we explore an algorithm to create annotations for SKE. 


\subsection{Temporal Alignment}
In frame-level feature representation, the video similarity is usually computed based on frame-to-frame similarities by taking into account the temporal sequence of frames. A simple method is to vote temporally by Temporal Hough Voting \cite{voting1,vcdb}. The graph-based Temporal Network (TN) \cite{temporaNetwork} takes matched frames as nodes and similarities between frames as weights of links to construct a network. And the matched clip is the weighted longest path in the network. Another method is Dynamic programming \cite{dp1} based on frame-to-frame matrix to find a diagonal blocks with the largest similarity. These methods are not capable of capturing great varying temporal similarity patterns \cite{kordopatis2019visil}. For example, a speed 2$\times$ transformation may cause a discrete path in TN network and a change of the slope of diagonal block in DP, which may result in a failure to detect path or pattern. 

ViSiL \cite{kordopatis2019visil} learns the similarity patterns between video pairs with heavy computational cost\cite{Shao2020ContextEF} but cannot locate segments. The approaches of \cite{zhou2020generating, Chen2020RethinkingTB} both introduce graph convolution to capture the temporal information between frames. Inspired by temporal matching kernel~\cite{Poullot2015}, \cite{LAMVBaraldi2018} transforms the kernel into a differentiable layer to find temporal alignments. \cite{feng2018video} proposes a cross-gated bilinear matching module and classify each frame into different labels. Although these two methods achieve decent results, both require to fuse pairwise video information heavily, which limits them from large-scale video retrieval. \cite{Hu_2019} formulates temporal alignment as an object detection task on the frame-to-frame similarity matrix, which is the most relevant work to our SPD. Different from \cite{Hu_2019}, our SPD is so designed that its loss can be backpropagated to SKE, leading to an end-to-end training in SSAN.

\section{Segment Similarity and Alignment}

The main contributions of our large-scale video retrieval system can be decomposed into four main parts: (i) Self-supervised keyframe Extraction (SKE), (ii)Similarity Pattern Detection (SPD) for alignment, (iii) Multi-task end-to-end joint learning of Segment Similarity and Alignment Network (SSAN) (iv) Index and retrieval with SSAN. This section describes the above parts in detail.




\subsection{Self-supervised Keyframe Extraction}
Capturing long-range temporal information is of great importance for video retrieval, especially for S-CBVR systems whose goal is not only to retrieve similar videos, but also to locate similar segments precisely. Therefore, the understanding of the whole video is indispensable. Moreover, the heavy computation cost may impose too great a burden on a large-scale CBVR system.
To tackle this issue, we formulate the keyframe extraction as a multi-frame classification problem, which can be inserted into the pipeline as shown in Figure 1. We adopt 2D temporal arrangement of frames to capture long-range representations. Furthermore, we investigate a lightweight classification network, which avoids explicit extraction of frame-level features,in order to reduce computational overhead.
 
 The detail process of SKE is shown in Figure 2. Firstly, we extract $n$ frames per second (noted as ${n}$ fps) uniformly as the basis frames $F$={($f_1$,...,$f_i$,...)} for each video. 
 Secondly,the basis frames are resized to small blocks with width and height equal to $s$ ($s$=32) and organized in a tiled pattern with $m$$\times$$m$ ($m$=24) blocks as \cite{hand2020}. This results in a sequence of tiled images with fixed size of [$s$$\times$$m$,$s$$\times$$m$]. 
 This 2D temporal arrangement of frames contributes to enlarge the receptive field. Specifically, given a frame $f_i$, we investigate the temporal information not only with fixed range [$i$-$k$/2, $i$+$k$/2] horizontally, where $k$=1,2,..,$m$, but also with the long-range dependencies [$i$-$l$/2*$m$, $i$+$l$/2*$m$] in vertical direction, where $l$=1,2,...,$m$.
 The bounding box location of a frame $f_i$ is
\begin{equation}
\label{eq1}
bbox_{GT}=[(i/m)*s,(i/m)*s,(i/m+1)*s,(i/m+1)*s]
\end{equation}

Thirdly, the tiled images are fed to a lightweight classification network to get the keyframe confidence score of each basis frame. Specifically, given a tiled image with $m$$\times$$m$ blocks, the classification network assigns the keyframe confidence score to each block, resulting in a probability matrix with size $m$$\times$$m$. Finally, the probability matrix is reshaped to a vector with size 1$\times$($m$$\times$$m$) and then the probability vectors of all tiled images are concatenated in the sequence of the original basis frames $F$={($f_1$,...,$f_i$,...)} for each video. 

The classification network is trained in a supervised way. In order to avoid the subjectivity and inefficiency of human annotations,
we explore a high performance and high computational complexity keyframe extraction method to generate annotations automatically, using frame-to-frame similarity matrix and dynamic programming scheme inspired by \cite{optimalgroup2016}.
In detail, we extract frame-level CNN features using an ImageNet pretrained mobilenet model and construct a frame-to-frame similarity matrix. Instead of calculating a normalized cost in \cite{optimalgroup2016}, we select keyframes by comparing similarities between neighborhood frames once they meet the conditions of similarity threshold and time limit.
In this way we can locate keyframes with significant variances precisely, as the frame-level features incorporate useful semantic information and the similarity matrix takes temporal structure into account.
The frames with poor quality or little information are classified as non-keyframes by pre-processing. Moreover, the generated keyframes satisfy requirements in capturing content variations in the retrieval problem.

\begin{figure}[ht!] 
  \centering
  \includegraphics[width=1\linewidth]{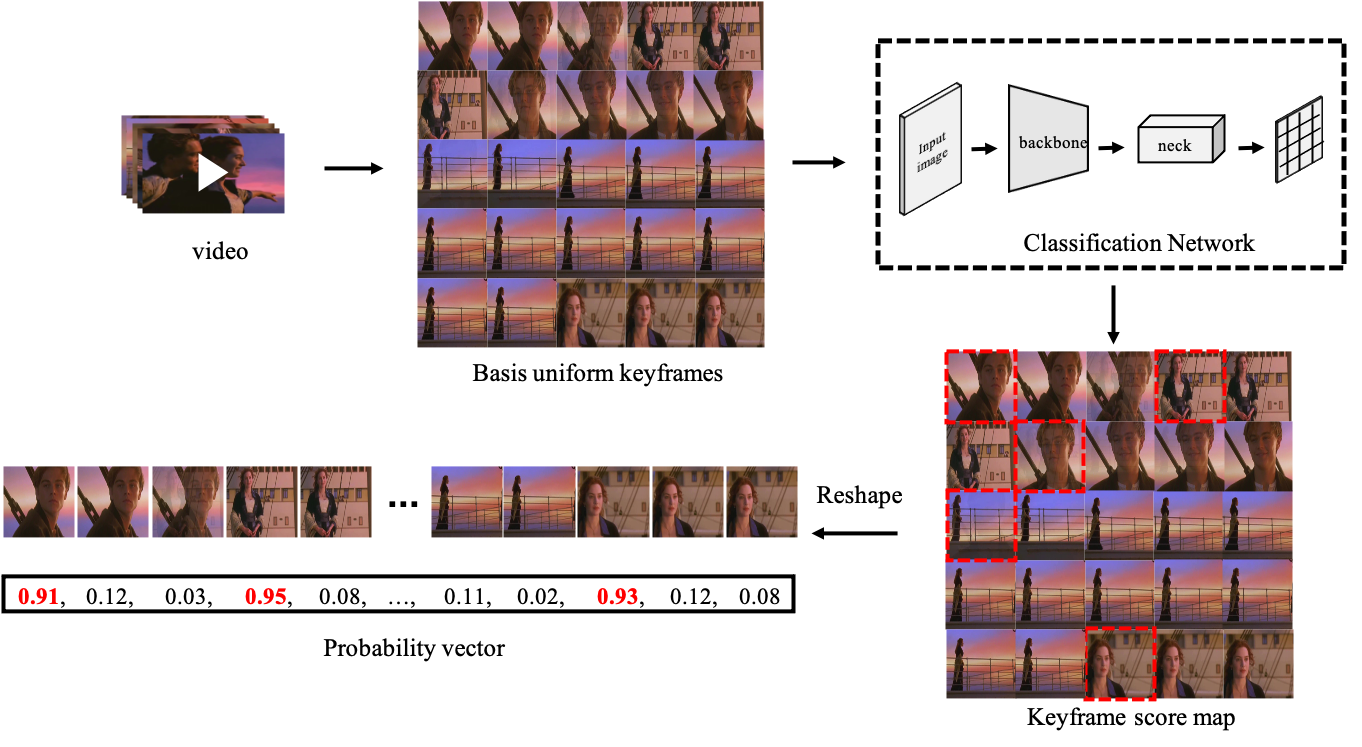}
  \caption{Self-supervised Keyframe Extraction (SKE) module}
\end{figure}

The training loss is defined as:
\begin{equation}
\label{eq2}
L_{SKE}=L_{BCEloss}
\end{equation}
where $L_{BCEloss}$ is a Binary Cross Entropy (BCE) classification loss on the category of each block (keyframe or not). The keyframe confidence score is $P_x=\frac{1}{1+\exp{-f(x)}}$ where $f(x)$ is the output of classification network. 
The objective is to extract the most representative and diverse frames in consideration of long-range representations and content variations.

In practice, there are some scenarios that the video frames are almost still with little changes, e.g., news reports or sports broadcasts. In these cases, SKE module only extracts 
very few frames in a relatively large compression rate, and this cannot provide enough information for the next-step process. Therefore, we interpolate the frame index to make sure that at least one frame is extracted within a sparse frame interval, and this is referred to as the sparse uniform mask in Figure 4. 

\subsection{Temporal Alignment based on Similarity Pattern Detection}
The frame-to-frame similarity matrix based on frame-level features is capable of representing temporal structure between the relevant videos and it has proved to be successful to learn similarity patterns of the pairwise frame similarities \cite{kordopatis2019visil}. However, ViSiL \cite{kordopatis2019visil} can only retrieve relevant videos, but fails to locate the exact temporal location of relevant segments. To solve this problem, we define alignment as a detection problem that detects similarity pattern on frame-to-frame similarity 
matrix $S$, as object detection is the coupling of object classification and localization. 
The matrix element $s_{ij}$ is the similarity between a frame feature $f_i$ of query video and $f'_j$ of a database video to be compared. 
\begin{equation}
\label{eq4}
  s_{ij}= f_i * f'_j/(||f_i||*||f'_j||)
\end{equation}
where $i$,$j$ indicates the $i$th frame of query video and the $j$th frame of the database video.

\begin{figure}[ht!] 
  \centering
  \includegraphics[width=1\linewidth]{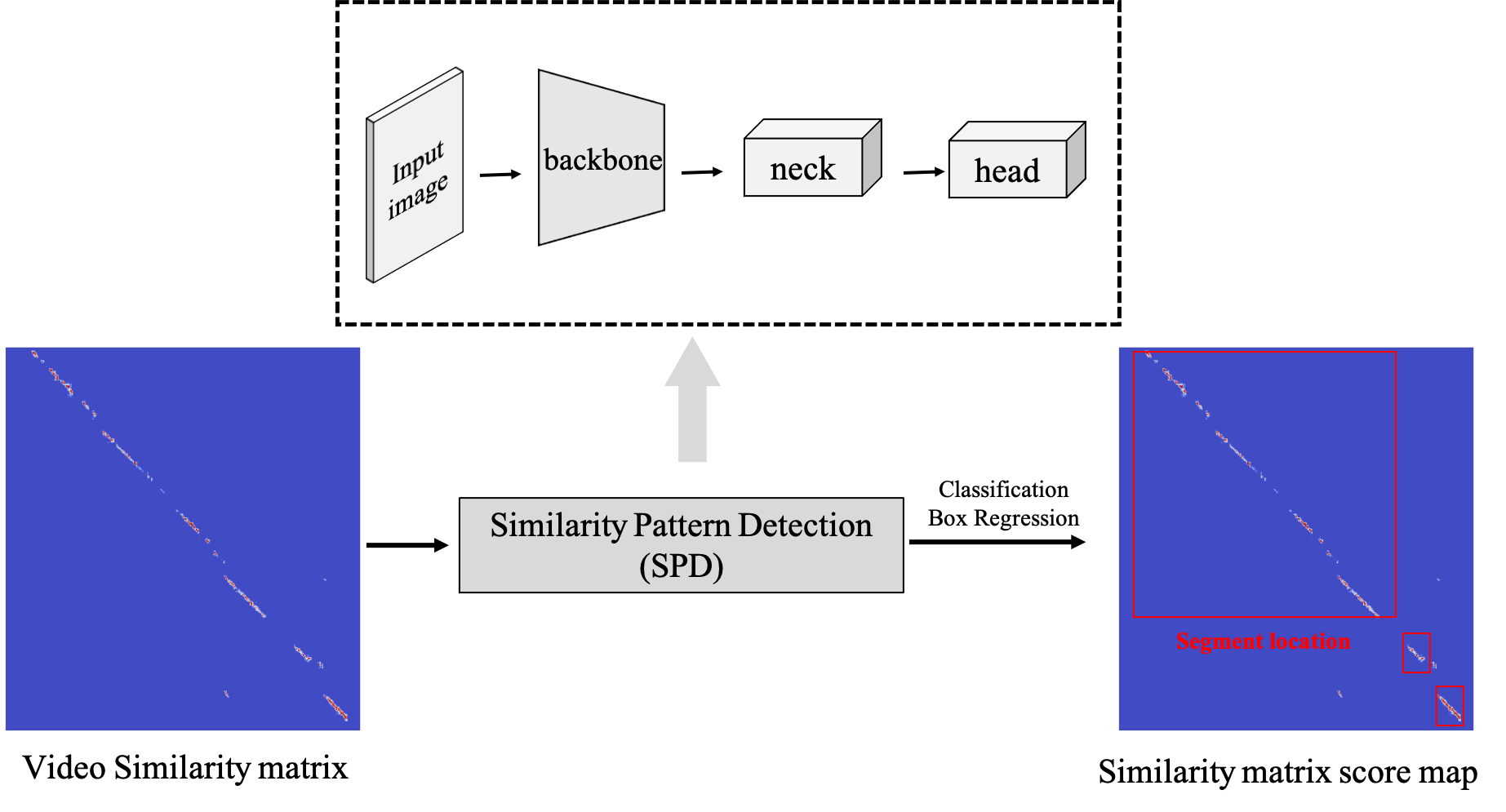}
  \caption{Similarity Pattern Detection (SPD) module}
\end{figure}

The ground truth bounding boxes $[t_i,t_j,t_i',t_j']$ are generated from annotated similar segments $[t_i,t_i']$ and $[t_j,t_j']$. The score of detected bounding box during inference indicates the confidence of the pattern and the location indicates the start and end time of similar segment in corresponding videos.The training loss is defined as:
\begin{equation}
\label{eq5}
L_{SPD}=L_{BCEloss}+L_{GIoU}
\end{equation}

where $L_{BCEloss}$ is the similarity pattern classification loss, which aims to predict two segments are similar or not, and $L_{GIoU}$ \cite{giou_2018_CVPR} is the bounding box location or similarity pattern location regression loss, whose objective is to learn the start and end time of similar segments. For multiple similar segments existing in two corresponding videos, there will be more than one bounding boxes detected and the score of the highest box is taken as the video similarity. 

\subsection{End-to-End and Multi-Task Joint Learning }

\begin{figure*}
\begin{center}
\includegraphics[width=14cm,keepaspectratio]{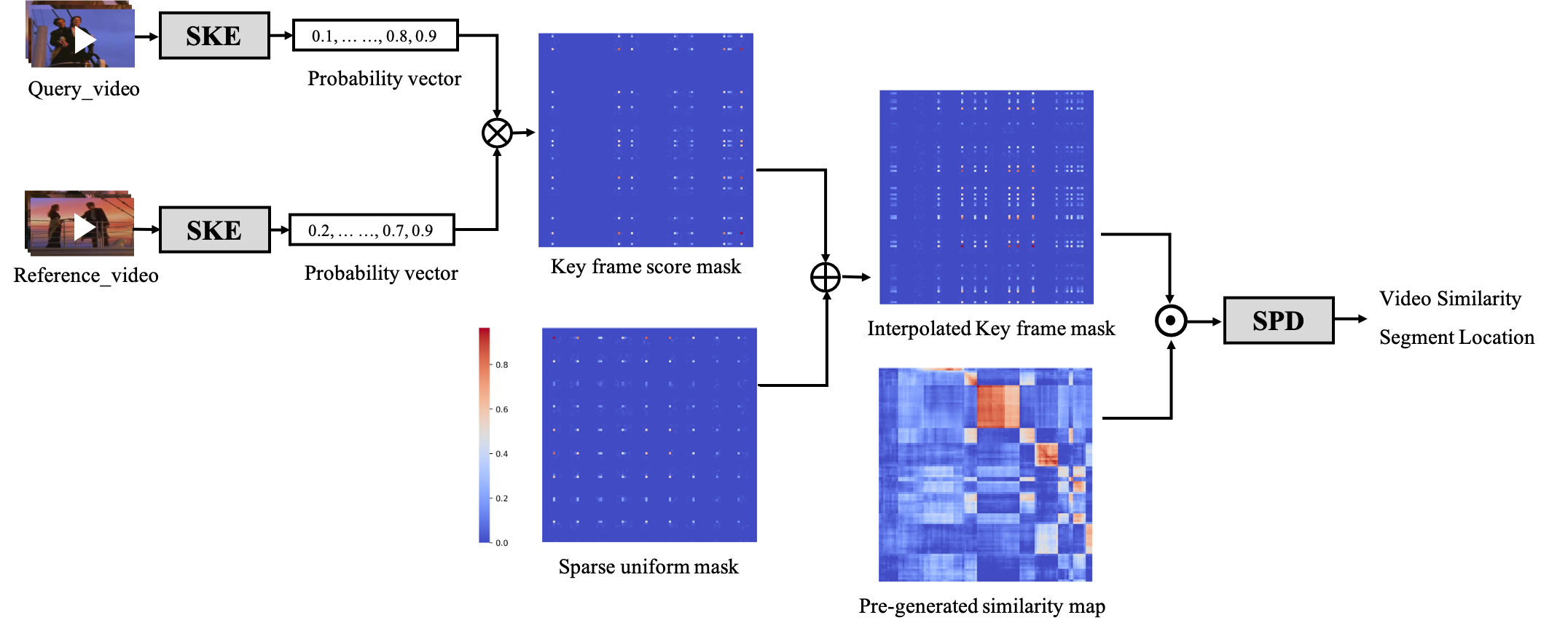}
\end{center}
   \caption{The training process of SSAN. The bottom similarity map is pre-computed in advance from features of a video pair (illustrated in Eq.(3)), which is different with index search part in query process of SSAN in Figure 1.}
\label{fig:short}
\end{figure*}
Due to the variability in video lengths, which vary from just a few seconds to several minutes/hours, and the diversity of similar segments, which can appear only partially in long videos, it is intuitive and beneficial to extract the most relevant frames that can represent the similarity pattern. Also, the introduction of keyframes makes it challenging to detect similar segments due to its variations in similarity patterns.
Since the SPD module is robust enough to capture various similarity patterns to a certain extent, we propose to integrate SKE and SPD in one network to extract keyframes and locate similar segments at the same time. Benefit from the end-to-end manner, we expect to focus on the necessary frames for preserving similarity pattern, and learn the variations in similarity pattern.
During training,minimizing objective function of SPD will guarantee the final temporal alignment accuracy.



Figure 4 illustrates our approach in training process. 
A video pair is fed to the network to construct the similarity matrix. We sample basis frames uniformly and densely from two videos. For one thing, we input them into our SKE to obtain 1D keyframe score vector $P^{v}_x$ =[$\nu_1$,...$\nu_n$] ($n$ is the number of basis frames) for each video,   
which is further fused to a pairwise 2D keyframe score mask. For another, the frame-level features of basis frames are extracted before training so that the frame-to-frame similarity matrix on basis frames $S$ is fixed. During training, the dense similarity matrix $S$ is multiplied by 2D keyframe score mask and then SPD learns how to detect similar patterns based on the weighted similarity matrix.

In order to get a high-quality keyframe distribution with adequate temporal information, the keyframe score mask based on SKE is fused with a sparse uniform mask. 
With the fusion of a keyframe score mask and a uniform similarity map $S$, we integrate SKE into SPD.
The total process is differentiable and can be trained in an end-to-end way.
\begin{equation}
\label{eq7}
SPD_{input} = P^{v_1}_x * P^{v_2}_x * S
\end{equation}
where $SPD_{input}$ is the input of temporal alignment in the joint network and $P^{v_1}_x$ and $P^{v_2}_x$ are keyframe score vectors of query video and database video respectively.

The total training loss is defined as:
\begin{equation}
\label{eq8}
L_{SSAN}=L_{SKE}+L_{SPD}
\end{equation}

During inference, a direct way to build keyframe similarity matrix is to extract features of keyframes and only calculate similarities between keyframes in the matrix as follows: 
\begin{equation}
\label{eq6}
m_{ij}=\left\{
\begin{aligned}
s_{ij}, \ & \text{if }  f_i, f'_j \in \text{ \{keyframes\}} \,; \\
0 , \ & \text{otherwise} \,.
\end{aligned}
\right.
\end{equation}

\subsection{Extend to Large-Scale Retrieval}
During training two videos are fed to the SSAN network to compute similarity matrix which will impede our method from applying to large-scale retrieval. Therefore we use an effective online inferring strategy based on index search. 
Firstly, SKE extracts keyframes by quantizing classification scores to 0,1 with a threshold. For database videos, only keyframe features are stored and indexed offline. For query, keyframe features are computed to search index. 
Secondly, the similarity matrices are built according to index search results.

In detail, the indexes of all the keyframe features of gallery videos are built in advance by high-dimensional indexing techniques. And the similarity matrix is constructed by similarity scores between the query and index search results. For a frame $f_i$ extracted from a query video by SKE ($k$ frames in total), we select top$N$ retrieved frames which are distributed in $M_i$ videos. Therefore, up to $\sum_{i=1}^{k}M_i$ video pairs are generated and then the results are merged to generate a single similarity map for each individual video pair. The whole process is shown in Figure 5. Compared with one vs one video matching, index-based similarity method avoids brute force comparison through all the gallery videos, which is essential for large-scale applications. Each similarity matrix is fed to the temporal alignment model (SPD) and inferred to get the score and location of similar segments.

\begin{figure}[ht!] 
  \centering
  \includegraphics[width=1\linewidth]{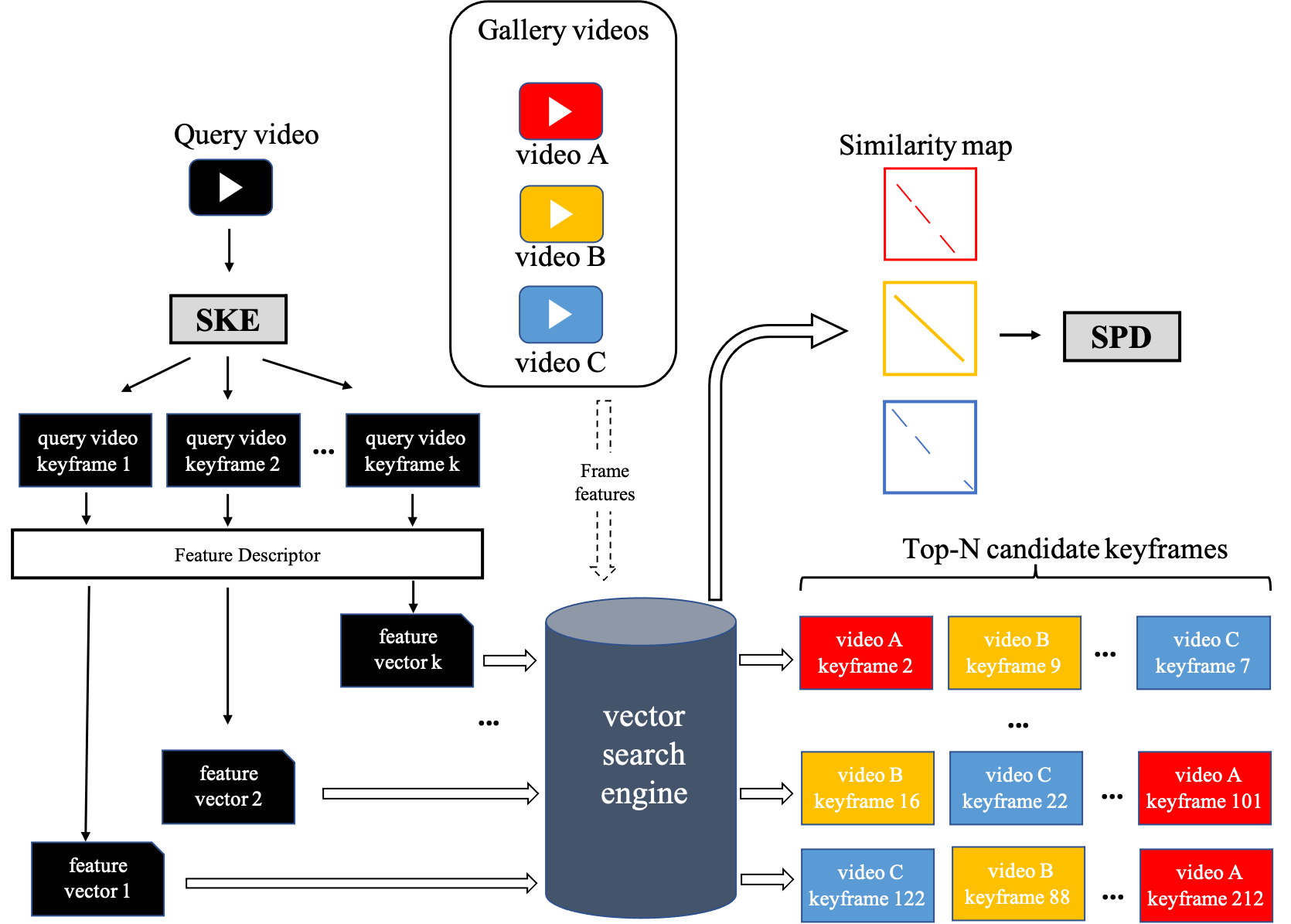}
  \caption{The pipeline of our segment-level video search using index in detail. The frame features of gallery videos are extracted and indexed offline (marked with dashed arrow). }
\end{figure}

\section{Experiments}
In this section, we present the datasets and evaluation metrics (Sect. 4.1), implementation details (Sect. 4.2), and the experimental results of our proposed approach (Sect. 4.3).
\subsection{Datasets and Evaluation Metrics}

\textbf{VCDB} \cite{vcdb} consists of 28 query sets and 528 videos with 9,000 pairs of copied segments in the core dataset.
The annotation gives the precise temporal location of each copied pair and thus is used for temporal alignment evaluation. 
Since VCDB is the only public dataset for VCD with complete segment-level annotations, we split the original VCDB into training set VCDB\_{train} and test set VCDB\_{test} to evaluate segment retrieval performance. 

Additionally, we further augment the VCDB\_{train} with some transformations of regular video editing as follows: a) $temporal$, including clipping, concatenating, accelerating, decelerating, dropping frames, or frame rate change, and b) $spatial$, including cropping, flipping, affine transformation, adding noise, gray-scale, or logo/watermark insertion.

\textbf{CC\_WEB} \cite{wu2007practical} consists of 24 query sets and 13,129 videos. This dataset contains duplicate and near-duplicate videos from web video search, which are approximately identical videos close to the exact duplicate of each other, but different in file formats, encoding parameters and photometric variations. The 24 queries are designed to retrieve the most frequently viewed and top favorite videos from YouTube. Here, we use the entire dataset and the 'cleaned' version as \cite{kordopatis2019visil} to measure the mean Average Precision (mAP).

\textbf{FIVR-200k} \cite{kordopatis2019fivr} consists of 225,960 videos and 100 queries. This dataset collects fine-grained incident retrieved videos. It includes three retrieval tasks: a) the Duplicate Scene Video Retrieval (DSVR), b) the Complementary Scene Video Retrieval (CSVR), and c) the Incident Scene Video Retrieval (ISVR). We use FIVR-5K for DSVR as \cite{kordopatis2019visil}, which is a subset of FIVR-200K by selecting the 50 most difficult queries in the DSVR task.

\textbf{VCDB\_plus} We extend VCDB core dataset with Havic \cite{HAVIC} and FIVR-200k \cite{kordopatis2019fivr} as distractors for large-scale video retrieval experiment. HAVIC is for MED (Multimedia Event Detection), which is comprised of approximately 3,650-hour videos from user-generated various events videos. The existing FIVR-200k (some of the video links have expired) and Havic dataset include about 270k videos with over 32 million frames. The feature database or index contains frames from both VCDB full dataset and all distraction videos.

\textbf{SVD}\cite{JiangHLLLL2019svd} contains 500,000 short videos less than 60 seconds for NDVR task, for which there is no need to localize segments, and thus is not used in this paper. 

For the S-CBVR task, there are two important aspects to measure the performance, namely segment-level temporal alignment and video-level retrieval ranking. We report F1 score for segment-level performance and mean Average Precision (mAP) for video-level retrieval task.
It is notable that only VCDB has segment-level annotations for F1 score evaluation. 
And the video retrieval performance is mainly evaluated on CC\_WEB with mAP. 
 
\subsection{Implementation Details}


\textbf{Feature Descriptor}
To depict frames, we employ R-MAC\cite{Tolias2016RMAC} and ViSiL\cite{kordopatis2019visil} features for different comparison experiments. The type of features has a great influence on retrieval preformance as \cite{kordopatis2019visil}. 
Here we extract R-MAC feature from the 29th activation map of a Resnet-34 with ImageNet pretrained model, which is consistent with LAMV \cite{LAMVBaraldi2018}. ViSiL feature is the $L_{3}$-iMAC feature as \cite{kordopatis2019visil}. 
R-MAC extracts a 512-dimensional vector and ViSiL outputs a 9$\times$3840-dimensional vector. 



\textbf{SKE and SPD training}
In the training process of SKE, the input video sequence is uniformly sampled with 8fps. 
We modified a lightweight network \cite{yolov5} to a classification task and train the network using SGD With Nesterov momentum 0.937, learning rate of 0.01 and weight decay of 0.0005. The model is trained for 200 epochs.
In the separate training process of SPD, we reshape the input similarity map with the shape of (640,640) and set the batch size to 32. We adopt the lightweight detection model \cite{yolov5} and mosaic \cite{YOLOv4} augmentation. 
We train the network using SGD parameters as SKE. The model is also trained for 200 epochs.

\textbf{End-to-End and Joint Training for SSAN}
To train SSAN, we input the tiled images with 24$\times$24 single blocks of a video pair into SKE, and then fuse the output 576-dimensional vector into a 576$\times$576 keyframe score matrix. The pre-computed uniform similarity map is also cropped and reshaped to (576,576).
The model is trained with a batch size of 8 using 8 GPU cards. 

%






\subsection{Experimental Results}
We demonstrate the performance of our proposed method on three tasks. Firstly, the segment-level alignment accuracy is evaluated on the VCDB. Secondly, we compare our video retrieval performance on the CC\_WEB. Finally, the effectiveness of our method on large-scale video retrieval task is demonstrated on the VCDB\_plus.

\subsubsection{\textbf{Temporal Alignment}}

We compare our proposed approach with existing methods\cite{Douze2015CTE,poullot:hal-TMK,LAMVBaraldi2018} and provide an ablation study to demonstrate the effectiveness of our proposed methods in the following sections.

\textbf{SPD} We first evaluate our SPD module with a uniform sampling of 8fps.
We run five-fold cross validation on VCDB.
For each fold, we take 23 video sets as the training set and the left 5 video sets as the test set. We test SPD with the 512-dimension R-MAC feature as  LAMV\cite{LAMVBaraldi2018} and the 9$\times$3840-dimension feature as ViSiL\cite{kordopatis2019visil}. The comparison results 
are shown in Table 1. Results of CTE/TMK/LAMV are obtained by open source codes provided in \cite{LAMVBaraldi2018}. Here, ($^*$) denotes evaluation on the whole VCDB dataset rather than the test set. Therefore, the result is slightly worse than that without ($^*$), which is consistent with the one reported in \cite{LAMVBaraldi2018}. It is shown that our proposed SPD based on both R-MAC and ViSiL features gives a higher F1 score than other methods with a lower frame rate in 
each fold. Note that fps in Table 1 is not execution time measurement, but a frame sampling rate. The lower the fps is, the less keyframes are extracted which further indicates less computational cost and storage. Although ViSiL feature brings the highest F1 score, its extremely high dimension is not feasible for large-scale video retrieval.
In the following experiments on VCDB, we will only report results on the first fold with the R-MAC feature due to space limitations. It should be noted that LAMV's F1 score drops from 77.61 to 68.70 when testing on the full dataset instead of the test set while SPD only slightly decreases from 81.86 to 80.39. Therefore, SPD is more stable when distraction videos increase, and this stability makes our method suitable for large-scale retrieval, which will be further introduced in Section 4.3.3 in detail. An example of temporal alignment between a video pair in VCDB is shown in Figure 6.

\begin{figure}[ht!] 
  \centering
  \includegraphics[width=1\linewidth]{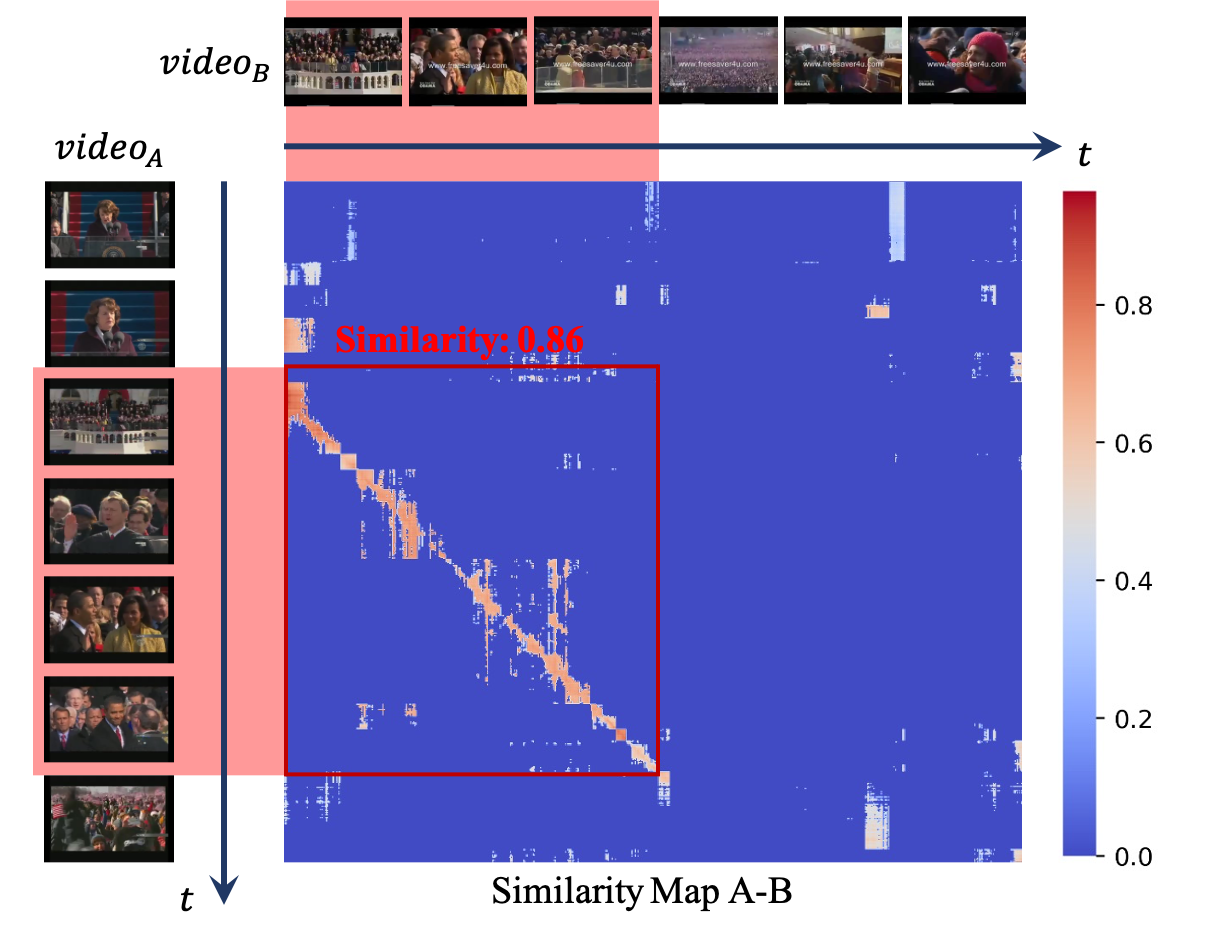}
  \caption{An alignment example on similarity map between a video pair of VCDB. The sampled frames from original videos with a red background indicate the similar segments, which are corresponding to the bounding box area in the similarity map.}
\end{figure}

\begin{table*}[]
\caption{Comparison of SPD and state-of-the-art for temporal alignment on VCDB with 5-fold evaluation. ($^*$) indicates video comparison range on full VCDB dataset while without ($^*$) on the testing subset.}
\begin{tabular}{@{}c|ccc|cccccc}
\toprule
\multirow{2}*{\textbf{Method}} &\multirow{2}*{\textbf{Extracted fps}}&\multirow{2}*{\textbf{Features}} &\multirow{2}*{\textbf{Dim.}}& \multicolumn{6} {c}{\textbf{F1 score}}\\
\cline{5-10}
&& & & \textbf{1st Fold} & \textbf{2nd Fold} & \textbf{3rd Fold} & \textbf{4th Fold} & \textbf{5th Fold} & \textbf{Average} \\ \hline

\textbf{CTE(m=16)}\cite{Douze2015CTE}  & 15 &  R-MAC  &  512  &  37.79             & 48.59             & 41.50              & 41.10              & 27.85             & 39.37            \\
\textbf{CTE(m=64)}\cite{Douze2015CTE}  & 15 &  R-MAC &  512 & 34.86             & 47.47             & 42.86             & 38.94             & 23.82             & 37.59            \\
\textbf{TMK}\cite{poullot:hal-TMK}        & 15 &  R-MAC  &  512 & 73.23             & 83.66             & 76.31             & 78.26             & 53.22             & 72.94           \\
\textbf{LAMV}\cite{LAMVBaraldi2018}      & 15 &  R-MAC  &  512  & 74.34             & 86.78             & 84.47             & 82.16             & 60.29             & 77.61            \\
\textbf{LAMV$^*$}\cite{LAMVBaraldi2018}      & 15 &  R-MAC  &  512  & 58.44             & 77.60             & 76.19             & 67.07             & 51.64             & 68.70            \\
\textbf{SPD(ours)}  & 8 &  ViSiL  &  9$\times $3840 & 81.34             & 86.90             & 90.54             & 86.20              & 80.95             & \textbf{85.19}            \\
\textbf{SPD(ours)} & 8 &  R-MAC &  512  & 80.27             & 78.69             & 85.53             & 84.03             & 80.80             & \textbf{81.86}   \\
\textbf{SPD$^*$(ours)} & 8 &  R-MAC &  512  & 79.41             & 78.46             & 83.42             & 80.13             & 80.55             & 80.39 \\ \bottomrule
\end{tabular}
\end{table*}

\textbf{SKE} The computational cost must be taken into consideration in practice, thus a lower frame sampling rate in Table 1 is more favorable. We conduct experiments in lower frame sampling rate regime to compare the uniform sampling strategy and the proposed SKE module where both are followed by SPD. The detailed results are shown in Table 2. With uniform frame rate, the temporal alignment results of 2fps and 1fps, i.e., the compression ratio of 25\% and 12.5\% slightly decline compared with origin 8fps reported in Section 4.3.1. 
We use the 8fps basis frames as SKE input and then obtain the keyframe index. The confidence score threshold in SKE is selected as 0.1 and 0.5 and the interpolation interval on the basis frame $f_i$ is 8 and 12, resulting in 19.81\% and 10.24\% of compression rate. The performance of the SKE module is slightly higher than the uniform sampling strategy with a even lower compression rate.

\textbf{SSAN} With the improved computational efficiency from SKE and alignment accuracy from SPD, we employ SSAN to achieve an end-to-end improvement. In the detailed implementation, the pretrained weight of SPD is obtained from the model based on 8fps uniform sampling (the second row of Table 2) and the pretrained weight of SKE is consistent with Section 4.3.2. After joint training of SSAN network, the segment-level performance is significantly improved by 2-3\% in comparable compression ratio. The detailed result is shown in the last row of Table 2.

Besides the improved alignment accuracy, another advantage of SPD is that multiple copied segments can be effectively detected by multiple bounding boxes, while LAMV can only output a global temporal offset between two videos.









\begin{table}[]
\caption{Uniform/SKE with SPD and SSAN based temporal alignment comparison on the first fold of VCDB. All the results utilize R-MAC feature with 512-dimension.}
    \centering
    \begin{tabular}{c|c|c|c}
    \toprule
\textbf{Method}  & \textbf{Compression Ratio} & \textbf{fps} & \textbf{F1 score} \\ \toprule
\textbf{LAMV}\cite{LAMVBaraldi2018}    & 1.8750                      & 15.00           & 74.34             \\ 
\textbf{uniform+SPD}     & 1.0000                  & 8.00            & 80.27             \\\midrule 
\textbf{uniform+SPD}     & 0.2500                       & 2.00            & 73.71             \\ 
\textbf{SKE+SPD} & 0.1981                     & 1.58         & 76.54             \\ \midrule 
\textbf{uniform+SPD}     & 0.1250                      & 1.00            & 71.59                  \\ 
\textbf{SKE+SPD} & 0.1024                     & 0.82         & 75.48             \\ \midrule 
\textbf{SSAN}     & 0.1432                      & 1.14            & \textbf{77.93}                  \\  \bottomrule
\end{tabular}
    \label{tab:alignment}
\end{table}

\begin{table}[]
    \centering
    \caption{The mAP results on CC\_Web retrieval. ($^*$) denotes evaluation on the entire dataset. }
    \begin{tabular}{c|c|c|c}
        \toprule
        \textbf{Method} & \textbf{support alignment} & \textbf{cc\_web} & \textbf{{cc\_web}$^*$}  \\ \hline
        \textbf{ViSiL}\cite{kordopatis2019visil} & $no$  & $0.996$ & $0.993$\\ \hline
        \textbf{TN}\cite{temporaNetwork} & $yes$ & $0.991$ & $0.987$ \\ \hline
        \textbf{DP}\cite{dp1} & $yes$  & $0.990$ & $0.982$ \\ \hline
        \textbf{SPD} & $yes$  & $0.992$ & $0.985$ \\  \bottomrule
    \end{tabular}
    \label{tab:retrival_ccweb}
\end{table}


\subsubsection{\textbf{Video Retrieval}}
We compare our video-level retrieval performance with the publicly available implementation of TN, DP and ViSiL \cite{kordopatis2019visil}. 
As mentioned in Section 4.1, our evaluation is built on CC\_WEB and FIVR. 
The training dataset is the original VCDB and VCDB\_train mentioned in Section 4.1. The detailed results of SPD and SSAN on CC\_WEB are shown in Table 3. It is shown that both SPD and SSAN achieve comparable video retrieval results with existing methods. Note that the ViSil with the highest mAP fails to output accurate temporal locations while other methods in Table 3 have alignment ability. 
For FIVR task, SPD and SSAN can only achieve mAP results of 0.74 and 0.75, both of which are lower than ViSiL. From close observation and case analysis on FIVR, we find that some very short segments of video A are interlaced and inserted into video B resulting in extremely small and irregular patterns in the similarity map, which can be further formulated as a small object detection problem. Due to lack of such types of transformation data in our training data, it is difficult to retrieve such videos. 

\subsubsection{\textbf{Large-Scale Alignment and Retrieval Performance}}

In this section, we discuss the results of large-scale experiments. We simulate the real scenario of video retrieval on the VCDB\_plus with 270k distraction videos. As the ViSiL feature is impractical for a large-scale retrieval system, we only adopt the R-MAC feature here. We run 5-fold cross-validation on the VCDB core dataset while fixing the distractors in the gallery and plot the averaged precision-recall curve as \cite{vcdb}. Figure.7 shows that the curve of our SPD with 8fps forms an upper envelope of the ones in \cite{vcdb}. The maximum F1 score of SPD with 270k distraction videos is 66.72\%, which is 6.4\% higher than the previous results in \cite{vcdb} without any distractors.

\begin{figure}[ht!] 
  \centering
  \includegraphics[width=0.9\linewidth]{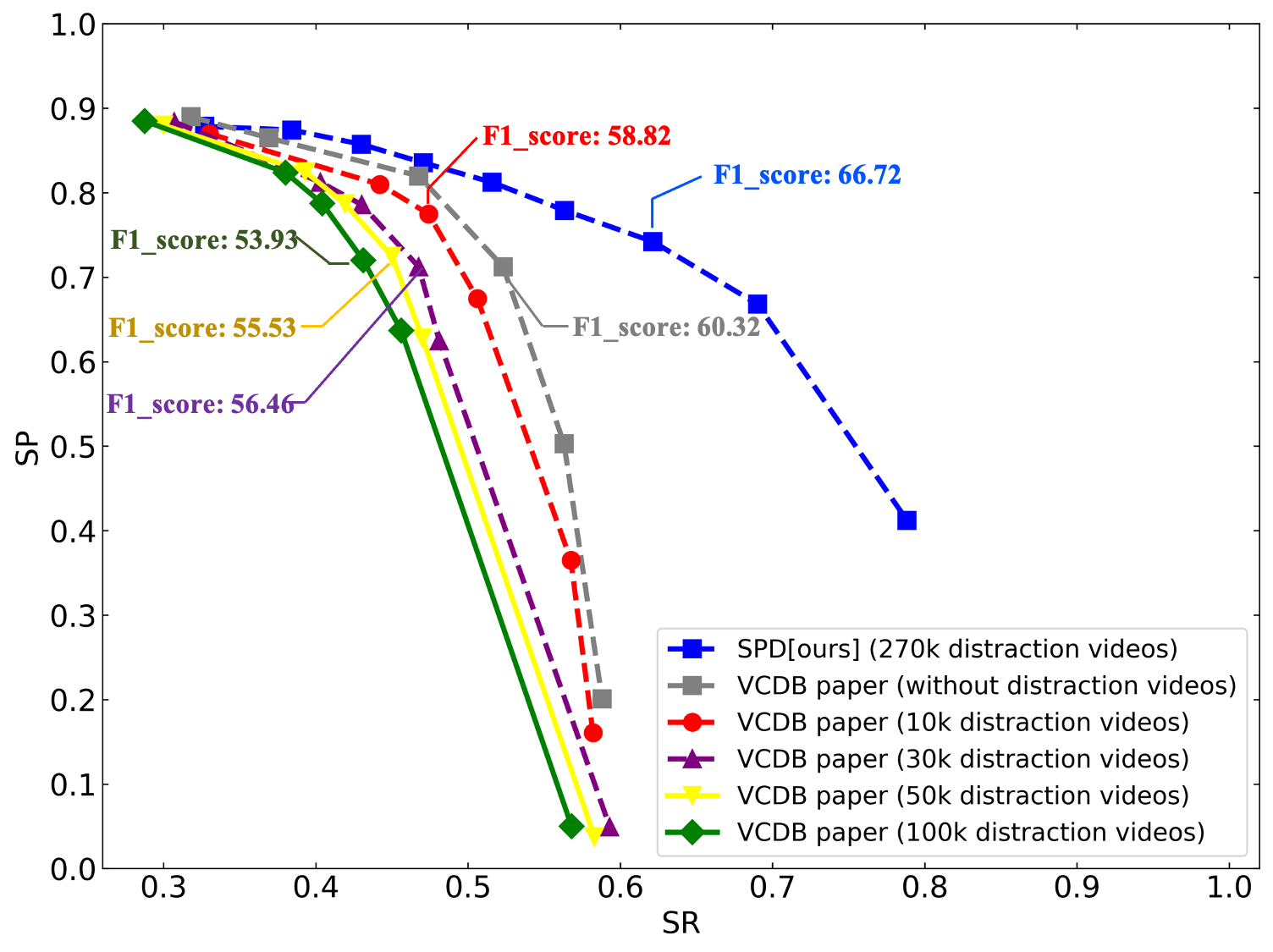}
  \caption{Precision-recall curves of large-scale video retrieval on VCDB from Ref \cite{vcdb} and our proposed SPD. We also denote the highest F1 score on each curve.
}
\end{figure}

Furthermore, we evaluate SSAN on the first fold of VCDB\_plus. The interpolation interval on 8fps basis frames is set as 4, leading to the compression ratio of 0.285. The results are presented in Table 4. Even with a large number of distractors and a lower frame sampling rate, SSAN still outperforms both LAMV and SPD with uniform sampling. With ~6.5 times fewer features than LAMV, SSAN gains 2.78\% improvement on F1 score evaluated on VCDB\_plus. The results demonstrate both the high alignment accuracy and high efficiency of SSAN.

It is also remarkable that the SKE module of SSAN not only saves the feature storage in the search engine but also reduces the query time cost as fewer keyframe features need to be computed and queried. The time cost for different stages of online query on the VCDB\_plus is about 0.043h, 0.056h and 0.250h for the SKE module, the index search and the SPD module respectively, resulting in a total time cost of 0.35h. As a comparison, the time cost of LAMV on the VCDB with the open-source code \cite{LAMVBaraldi2018} is over 5 hours. All the above experiments are performed on Intel(R) Xeon(R) Platinum 8163 CPUs and Nvidia V-100 GPUs.

\begin{table}[]
    \centering
    \caption{Evaluation of large-scale segment-level retrieval on first fold of VCDB.}
    \begin{tabular}{c|c|c|c}
         \toprule
        \textbf{Method} & \textbf{Distraction video number} & \textbf{fps}  & \textbf{F1 score}  \\ \hline
        \textbf{LAMV}\cite{LAMVBaraldi2018} & 0 & 15.00  & 58.44 \\ \hline
        \textbf{SPD} & 270k & 8.00  & 60.04 \\ \hline
        \textbf{SSAN} & 270k & \textbf{2.28}  & \textbf{61.22}  \\ \bottomrule
    \end{tabular}
    \label{tab:retrival_fivr}
\end{table}


 
\section{Conclusion}
In this paper, we propose a Segment Similarity and Alignment Network (SSAN) for large-scale S-CBVR system, which integrate a Self-supervised Keyframe Extraction (SKE) module and a robust Similarity Pattern Detection (SPD) module to an end-to-end and multi-task joint learning network.

The SSAN 
can extract high-quality keyframes and conduct temporal alignment robustly.
Consequently, it achieves high alignment accuracy at a low computational cost and storage space.
In the furture, we will apply deep hash techniques on SSAN to further improve the storage consumption and search time latency in large-scale retrieval.


\bibliographystyle{ACM-Reference-Format}
\balance
\bibliography{sample-base}

\end{document}